\documentclass{article}
\usepackage[noend]{algorithmic}
\usepackage{microtype}
\usepackage{graphicx}
\usepackage{xspace}
\usepackage{enumitem}
\usepackage{eufrak}
\usepackage{subcaption}
\usepackage{multirow}
\usepackage{pifont}
\usepackage[normalem]{ulem}
\useunder{\uline}{\ul}{}
\usepackage{hyperref}

\usepackage[accepted]{icml2024}

\usepackage{amsmath}
\usepackage{amssymb}
\usepackage{mathtools}
\usepackage{amsthm}

\usepackage[capitalize,noabbrev]{cleveref}

\theoremstyle{plain}

\theoremstyle{definition}

\theoremstyle{remark}

\usepackage[textsize=tiny]{todonotes}
\usepackage{enumerate}
\usepackage{makecell}
\usepackage{adjustbox}
\usepackage{dsfont}
\usepackage{wrapfig}
\usepackage{xspace}
\usepackage{enumitem}
\usepackage{eufrak}
\usepackage{xcolor}
\usepackage[hang,flushmargin]{footmisc}

\newcommand{\SysName}{\texttt{PRIME}\xspace}

\newcommand{\DataName}{\texttt{VIOLENT}\xspace}

\icmltitlerunning{PRIME: Protect Your Videos From Malicious Editing}

\begin{document}

\twocolumn[
\icmltitle{PRIME: Protect Your Videos From Malicious Editing}




\begin{icmlauthorlist}
\icmlauthor{Guanlin Li}{yyy,comp}
\icmlauthor{Shuai Yang}{yyy,comp}
\icmlauthor{Jie Zhang}{yyy}
\icmlauthor{Tianwei Zhang}{yyy}
\end{icmlauthorlist}

\icmlaffiliation{yyy}{Nanyang Technological University}
\icmlaffiliation{comp}{S-Lab}
\icmlcorrespondingauthor{Guanlin Li}{guanlin001@e.ntu.edu.sg}


\vskip 0.3in
]



\printAffiliationsAndNotice{}

\begin{abstract}
With the development of generative models, the quality of generated content keeps increasing. Recently, open-source models have made it surprisingly easy to manipulate and edit photos and videos, with just a few simple prompts. While these cutting-edge technologies have gained popularity, they have also given rise to concerns regarding the privacy and portrait rights of individuals. Malicious users can exploit these tools for deceptive or illegal purposes. Although some previous works focus on protecting photos against generative models, we find there are still gaps between protecting videos and images in the aspects of efficiency and effectiveness. Therefore, we introduce our protection method, \SysName, to significantly reduce the time cost and improve the protection performance. Moreover, to evaluate our proposed protection method, we consider both objective metrics and human subjective metrics. Our evaluation results indicate that \SysName only costs 8.3\% GPU hours of the cost of the previous state-of-the-art method and achieves better protection results on both human evaluation and objective metrics. Code can be found in \url{https://github.com/GuanlinLee/prime}.
\end{abstract}

\section{Introduction}


%
Recently, there has been a notable emergence of advanced generative models, including various diffusion models~\cite{ho_denoising_2020,song_denoising_2021,song_score-based_2021}, which have become readily available on public platforms like Hugging Face~\cite{HuggingFace} and Civitai~\cite{Civitai}.
%
%
%
These models exhibit the capability to generate realistic photos when provided with specific conditions or prompts. 
Beyond mere photo generation, there has been substantial research into the realm of photo editing, encompassing the addition of new elements or the alteration of styles. Notably, the latent diffusion model (LDM)~\cite{rombach_high-resolution_2022} has been instrumental in this pursuit.
%
Moreover, recent endeavors have turned to the domain of video editing~\cite{khachatryan_text2video-zero_2023,khandelwal_infusion_2023,qi_fatezero_2023}, which can be viewed as a sequence of images arranged along the temporal dimension.
%
These works aim to ensure the coherence and consistency of these stacked images, employing techniques such as global attention constraints~\cite{geyer_tokenflow_2024} and latent feature constraints~\cite{yang_rerender_2023,khachatryan_text2video-zero_2023}. 
%
Nevertheless, advanced video editing also opens the door to the potential creation of illegal or malicious videos. Therefore, it is critical to protect videos from malicious editing.





Several previous works have been proposed to safeguard images from unauthorized use or misuse 
by introducing adversarial perturbation in advance. 
One notable example is Photoguard~\cite{salman_raising_2023-1}, which effectively hinders the efforts of LDMs, compelling these models to generate poor editing outcomes for a given image.
Besides, some other research endeavors~\cite{shan2023glaze, le_anti-dreambooth_2023-1, rhodes_my_2023, zheng_understanding_2023} have been directed towards protecting specific artistic styles or objects from being utilized in training generative models.
While these methods have proven effective in protecting static images, we have identified shortcomings in extending these protections to videos.
First, video editing methods can cooperate with various LDMs, which could be fine-tuned using images from diverse sources unrelated to the frames within the protected videos. 
This invalidates previous protections~\cite{le_anti-dreambooth_2023-1,rhodes_my_2023,zheng_understanding_2023} that add perturbation to the protected targets, i.e., videos, to against fine-tuning. 
Second, previous methods of protecting images are time-consuming,
namely, adding perturbations to video frames would require hours for even brief video clips, making them impractical.
Third, the codec used in video coding will 
assign different compression ratios to individual frames
to achieve the best trade-off between the video quality and the file size. 
This dynamic compression applied to each frame will diminish the effect of perturbations applied to them, leading to failed protection.
The gaps mentioned above motivate us to explore and design new protection methods for videos.




To remedy gaps, we propose a novel protection method \SysName:  \texttt{PR}otect v\texttt{I}deos from \texttt{M}alicious \texttt{E}diting. 
According to the above shortcomings, we point out our design goals, i.e., 1) \textbf{zero-shot ability}, 2) \textbf{per-frame perturbing}, and 3) \textbf{anti dynamic compression}. 
Firstly, we shall guarantee \textbf{zero-shot ability}, because existing popular video editing methods directly can adopt various pre-trained models to launch the whole pipeline with arbitrary prompts. The protector has no information about the models, editing pipelines, and prompts. For this, \SysName protects videos based on the insight of the transferability~\cite{papernot_transferability_2016} of adversarial perturbation across various models and editing methods. To improve the transferability among different LDMs, we consider the latent features during the diffusion process and the final outputs.
Then, \textbf{per-frame perturbing} implies that every frame in the given video should be perturbed to ensure that the constraints, such as global attention, used in the video editing pipeline will not rectify the wrong features of the perturbed frames with the clean frames. Besides, per-frame perturbing will cost a lot of time and GPU resources. Therefore, we further propose two mechanisms, i.e., \textit{fast convergence searching} and \textit{early stage stopping}, to reduce consumption.
Finally, \textbf{anti dynamic compression} requires us to guarantee that the perturbation is robust to dynamic compression imposed by the video codec, making the generated protected videos inherit the properties of single frames. Usually, the compression algorithm is complex and lossy, making it impossible to maintain lossless perturbation after compression. Following an in-depth examination of existing video codec methods, we propose a straightforward yet highly effective approach to discretize perturbation within the pixel space, which can improve the robustness of perturbation during compression.

In the area of video editing, we observe that previous studies~\cite{geyer_tokenflow_2024, yang_rerender_2023, wu_fairy_2023} usually collect video clips sourced from the Internet to assess the effectiveness of their methods.
Similarly, we collect video clips to help us evaluate the performance of both video editing and protection techniques. To create malicious content, we propose two types of malicious editing tasks, i.e., the malicious NSFW (Not Safe For Work) editing task and the malicious swapping editing task. 
Due to the lack of ground-truth references, it is difficult to make a comprehensive and reasonable evaluation of the edited video. To address this, we consider the evaluation methods used in previous works~\cite{geyer_tokenflow_2024,yang_rerender_2023,wu_fairy_2023}, wherein we engage volunteers to participate in surveys gauging the quality and preferences of the video. 
Based on the human evaluation, we prove that existing video editing pipelines can produce high-quality videos for these two tasks, and obtain 2.99 out of 5 and 3.17 out of 5 for the video quality scores, respectively. On the other hand, the results prove that \SysName can significantly reduce the generated videos' quality (1.54 out of 5 and 2.44 out of 5 for two tasks, respectively). 
Overall, our contributions can be summarized as follows:
\begin{itemize}[leftmargin=*]
    \item We propose a new black-box video protection method, \SysName, against malicious video editing. We improve the transferability of our protection by simultaneously considering both latent codings and generated images. 
    
    \item \SysName is time-saving with our proposed \textit{fast convergence searching} and \textit{early stage stopping} mechanisms. \SysName only costs about 8.3\% GPU hours of the cost of Photoguard on the same hardware platform under the same protection budgets.
    \item \SysName can combat dynamic compression from the lossy video codec with our proposed anti compression method. \SysName increases about 8\% of the bitrate for protected videos, compared with protected videos from Photoguard. 
    
    \item Our results prove that \SysName has better protection performance and transferability than Photoguard. Only 18\% and 9\% of people think that the edited videos have better quality under the protection of \SysName than under the protection of Photoguard for two editing tasks, respectively.
\end{itemize}

\section{Related Works}

\subsection{Video Editing with LDMs}
A growing number of work~\cite{parmar_zero-shot_2023,wu_latent_2023,lin_mirrordiffusion_2024} focuses on turning a latent diffusion model (LDM)~\cite{rombach_high-resolution_2022} into a zero-shot image editor, and great progress is made. Such progress inspires the video editing area. Video editing is a subtask in the video generation area~\cite{blattmann_align_2023,chen_videocrafter1_2023,ge_preserve_2023}. Different from the general video generation task, which usually only requires a conditional prompt to guide the generation process, video editing requires a source video and a guidance prompt as conditions. When editing a given video, it can be seen as a series of images stacked along the time dimension. However, making the frames of the edited video consistent is still an open problem. Directly editing each frame of the given video will probably lead to different backgrounds or different poses for the foreground objects, due to the lack of pixel-level constrain. The recently proposed video editing frameworks are based on various pre-trained LDMs~\cite{rombach_high-resolution_2022}. To keep the consistency between frames, cross-frame global attention~\cite{geyer_tokenflow_2024,yang_rerender_2023,wu_fairy_2023} is widely used in these editing frameworks. On the other hand, some frameworks~\cite{yang_rerender_2023,khachatryan_text2video-zero_2023} adopt other conditions, such as depth, pose, and edge, to better enhance consistency. Considering that more and more advanced methods are proposed to generate high-quality videos, we believe there exists a potential risk that users can adopt these editing frameworks to modify videos for malicious purposes, making the edited ones illegal, misleading, or harmful. Therefore, we conduct the first research work in this area.


\subsection{Image Protection Against LDMs-based Misuse }
\vspace{-0.5em}
    Latent diffusion models (LDMs) can edit images based on conditional prompts, which can potentially be exploited to generate malicious content. To counter this threat, Photoguard~\cite{salman_raising_2023-1} has been introduced as a protective measure for images, aiming to hinder the efforts of LDMs. This method incorporates adversarial perturbations into images, effectively perplexing LDMs and preventing unauthorized editing.
Furthermore, LDMs can quickly learn specific objects or artistic styles by personalized techniques like DreamBooth~\cite{ruiz_dreambooth_2023}. 
To protect intellectual property or portrait rights, some works~\cite{shan2023glaze,le_anti-dreambooth_2023-1,rhodes_my_2023,zheng_understanding_2023} add perturbation into images before releasing them on the Internet. With such perturbed images, the fine-tuned LDMs are only capable of producing low-quality results.

In a nutshell, Photoguard focuses on protecting images from malicious editing in the inference stage of LDMs, while other techniques aim to impede the fine-tuning of LDMs. Considering that most video editing frameworks operate leveraging pre-trained models, we think that Photoguard stands as a baseline method for protecting videos similarly.


\section{Preliminary}
\vspace{-0.5em}
In this section, we aim to briefly introduce the LDMs and give an example to show how to edit photos with LDMs. Firstly, LDMs contain an image encoder $\mathcal{E}$, a U-Net $\mathcal{U}$, and an image decoder $\mathcal{D}$. The image encoder can project a given image $x$ to its latent code $f_\mathcal{E} = \mathcal{E}(x)$. The image decoder can project a latent code to an image $x = \mathcal{D}(\mathcal{E}(x))$. The U-Net $\mathcal{U}$ accepts the latent codes $f_\mathcal{E}$ and is related to a diffusion process, which contains a noise-adding forward process and a denoising sampling process. For the forward process, given a time series $t=[1, \dots, T]$, we have the following relation between clean $f_{\mathcal{E},0}$ and noisy $f^{d}_{\mathcal{E},t}$:

    $q(f^{d}_{\mathcal{E},t} | f_{\mathcal{E},0}) = \mathcal{N}(f^{d}_{\mathcal{E},t}; \sqrt{\bar{a}_t}f_{\mathcal{E},0}, (1-\bar{a}_t)\mathbf{I}), t = [1, \dots, T]$,

where $\mathcal{N}$ stands for a Gaussian distribution, $\bar{a}_t$ is a hyperparameter related to the diffusion process, $f_{\mathcal{E},0} = \mathcal{E}(x)$ is the clean latent, and $f^{d}_{\mathcal{E},t}$ is the noisy latent at time step $t$. For the sampling process, $f^{s}_{\mathcal{E},t-1}$ can be predicted with $\mathcal{U}(f^{s}_{\mathcal{E},t}, t, c_p)$ under DDIM sampling~\cite{song_denoising_2021}, where $c_p$ is a given condition based on a prompt and $f^{s}_{\mathcal{E},T} = f^{d}_{\mathcal{E},T}$ which is the boundary condition. 
For example, if we want to swap the face of a photo $x$ of Joe Biden with Donald Trump, we can first obtain the latent code $f_{\mathcal{E},0} = \mathcal{E}(x)$. Then we add noise to $f_{\mathcal{E},0}$ and obtain $f^{d}_{\mathcal{E},T_1}$, where $T_1 \le T$. With the prompt $c_p$ of ``a photo of Donald Trump'', we sample $f^{s}_{\mathcal{E},t-1}$ based on $\mathcal{U}(f^{s}_{\mathcal{E},t}, t, c_p)$ and $f^{s}_{\mathcal{E},T_1} = f^{d}_{\mathcal{E},T_1}$ step by step and get $f^{s}_{\mathcal{E},0}$, which is the clean latent for prompt $c_p$. Finally, we use $\mathcal{D}$ to convert $f^{s}_{\mathcal{E},0}$ to a photo $f_{\mathcal{D}}$, in which the edited face belongs to Donald Trump when $T_1$ is set properly. 

\section{Protect Videos from Malicious Editing}
\label{sec:protection}

In this section, we first clarify our motivation and the threat model. Then, we introduce the proposed \SysName in detail.

\subsection{Motivation}
\vspace{-0.5em}
An increasing number of users are sharing their video content on the Internet through popular social applications such as TikTok and Instagram.
However, this widespread sharing presents a substantial risk to the public's rights, as some users may exploit the capabilities of powerful LDMs for malicious purposes.
These malicious actions could include altering video content to make individuals appear unclothed or inserting weapons into the video, among other harmful editing.
Therefore, it is vital to explore how to protect videos from malicious editing effectively and efficiently.

\subsection{Threat Model}
\vspace{-0.5em}


\noindent\textbf{Attacker's Ability and Goal}.
%
The malicious user can directly download a pre-trained model from the public platform and then conduct malicious video editing. In this paper, we study two types of malicious video editing. The first type, called \underline{malicious NSFW editing}, incorporates ``not safe for work (NSFW)" elements into a given video, such as blood, drugs, and explicit nudity.
The second type, termed \underline{malicious swap editing}, centers around identity substitution within the given video, where one individual's identity is replaced with that of another. For example, replacing Joe Biden with Donald Trump in a video of the oath of office of the president.

\noindent\textbf{Protector's Ability and Goal}.
For the video protector, we assume a challenging black-box scenario, wherein the protector has no information about \textit{the editing models}, \textit{the editing methods}, and \textit{the attack types} that will be used by the malicious user. 
The video protector only has access to public pre-trained LDMs, which may not be the one that the malicious user uses, and adds perturbation to each frame before uploading the video to the Internet. The goal of the protector is to ensure that protected videos effectively thwart the efforts of malicious users, causing their results low-quality or preventing them from injecting NSFW content (such as scenes depicting violence or nudity) or engaging in identity swapping.


\subsection{\SysName}
\vspace{-0.5em}
As mentioned above, when
applying conventional image protection methods to safeguard videos, there are two weaknesses: 
the excessive time required and the vulnerability to the compression mechanisms utilized by video codecs.
To address it, we propose \SysName, having \textbf{zero-shot ability}, to improve the previous methods from two parts, i.e., \textbf{per-frame perturbing} and \textbf{anti dynamic compression}. We will first give a detailed introduction about each of them. Then, we will show how \SysName combines them.

\textbf{Zero-shot Ability}. With a pre-trained LDM $\mathcal{F}$, consisting of several parts such as an encoder $\mathcal{E}$, a decoder $\mathcal{D}$, a U-Net $\mathcal{U}$, the protector aims to generate adversarial perturbation for a given frame to cause $\mathcal{F}$ giving low-quality results. As the protector has no information about the models and video editing pipelines used by the malicious users, the generated perturbation should be generalizable for other LDMs. Furthermore, considering that the malicious users will adopt various prompts to edit the videos, the perturbation should be able to invalidate as many prompts as possible. 

To address such a challenge, we follow the previous method, Photoguard~\cite{salman_raising_2023-1}, to perform a diffusion attack, which means that \SysName considers not only $\mathcal{E}$, but also $\mathcal{U}$ and $\mathcal{D}$, i.e., the target is to find such a perturbation of the current frame, causing latent features disrupted and $\mathcal{D}(f^{s}_{\mathcal{E},0})$ close to a predefined target image.
Because the protector has no information about the specific editing pipeline employed by malicious users, we do not consider intricate technologies used in video editing methods, such as cross-frame attention and latent constraints, to make the protection more general. Similar to Photoguard, \SysName depends on the transferability of the perturbation and can be further cooperated with methods~\cite{kurakin_adversarial_2017,athalye_synthesizing_2018,song_physical_2018} to further improve the robustness of the perturbation.

\textbf{Per-frame Perturbing}. We find that in existing video editing pipelines~\cite{geyer_tokenflow_2024,yang_rerender_2023,khachatryan_text2video-zero_2023,wu_fairy_2023}, global attention constraints can rectify artifacts and
address imperfections in the structure level of videos. This correction is achieved by referencing information from other frames within the video. 
Therefore, if not all frames are protected, the generated video will still be maintained at a desirable level.
Such observation requires the protector to add perturbation to every frame. However, unlike images, protecting videos containing many frames will take an extremely long time. To reduce time consumption, we introduce two mechanisms, i.e., \textit{fast convergence searching} and \textit{early stage stopping}. 

In previous work, the perturbation $\delta$ is oriented to a fixed target image $\hat{x}$, which means the protector aims to make $\mathcal{F}$ generate a result closing to $\hat{x}$ for the protected image $x$, i.e., the optimization can be described as:

\centerline{ 
$
\min_\delta d( \hat{x}, \mathcal{F}(x + \delta)),
$}

where $d( \cdot, \cdot)$ gives the distance between inputs and $\mathcal{F}(\cdot)$ represents a complete diffusion process to generate images with $\mathcal{F}$. However, we observe that the convergence speed for a different clean image $x$ will be different when generating the perturbation. It will take more optimization steps to obtain perturbation for $x$ having a slow convergence speed. Therefore, we propose the \textit{fast convergence searching} to find a better target image for each frame. Specifically, given a video $V$, which is constructed using a sequence of frames $V = [x_1, x_2, \dots, x_n]$, we maintain a queue of potential target images $Q=\{\hat{x}_1, \hat{x}_2, \dots, \hat{x}_N\}$\footnote{We use the validation set of ImageNet~\cite{deng_imagenet_2009} in our experiments as $Q$.}. For each $x_i$, we select a target image from $Q$ to have the lowest similarity score $s_j$ based on the equation:
\vspace{-3pt}
\begin{equation}
s_j = \left\{
    \begin{aligned}
       &\mathrm{SIM}_{\mathcal{E}_{img}}(x_i, \hat{x}_j), \ \ \ \ \ \ \ \ \ \ \ \ \ \ \ \ \ \ \ \ \ \ \ \ \ \ \ \ \ \ \ \ \ \ \ \ \ \  i = 1\\
    &\mathrm{SIM}_{\mathcal{E}_{img}}(x_i, \hat{x}_j) + \mathrm{SIM}_{\mathcal{E}_{img}}(\hat{x}_{j_{i-1}}, \hat{x}_j),  \ i > 1
    \end{aligned}
    \right. \label{eq:1},
\end{equation}
and
\begin{align}
    \mathrm{SIM}_{\mathcal{E}}(&x_1, x_2) = \frac{\mathcal{E}(x_1)\otimes \mathcal{E}(x_2)}{\vert \mathcal{E}(x_1)\vert \cdot \vert\mathcal{E}(x_2)\vert} \nonumber \\
    j_i  &=  \mathrm{arg min}_j s_j, \hat{x}_j \in Q, \label{eq:2}
\end{align}
\vspace{-3pt}
\noindent where $\mathcal{E}_\mathrm{img}$ is the image encoder from the CLIP~\cite{radford_learning_2021}, $\vert \cdot \vert$ stands for the norm of the vector, $\otimes$ is the matrix multiplication, and $j_i$ is an index for images in $Q$. When $i=1$, $s_j$ only depends on the first term. Specifically, the second term is to ensure the perturbed continuous frames have different features, to better break the global attention constraints and increase flickers in the edited video. $\hat{x}_{j_{i-1}}$ is the target image for frame $x_{i-1}$. After selection, we obtain $\hat{x}_{j_{i}}$ for $x_i$. Because the target image and the corresponding frame are very different in the latent space, the convergence speed of the optimization process will be faster at the start.

Based on another observation that for a given perturbation budget $\epsilon$ under the $l_p$-norm, increasing the number of optimization iterations will only bring marginal improvement if the perturbation $\delta$ converges in several steps. It inspires us to introduce the \textit{early stage stopping} to further decrease the total time consumption. Specifically, we monitor the similarity $c_k^i$ between the latent generated by $\mathcal{E}$ for the current frame $x_i$ and all previous perturbed frames, i.e.,
\begin{align}
    c_k^i = \max_j \mathrm{SIM}_\mathcal{E}(x_i +\delta_k, x'_j)\label{eq:3},
\end{align}
where $\delta_k$ is the perturbation $\delta$ in the $k$-th optimization step, and $x'_j$ is the perturbed frame of $x_j$. When $c_k^i$ does not decrease, we will stop the optimization process and use the perturbation $\delta_k$ for $x_i$. Combining \textit{fast convergence searching} and \textit{early stage stopping}, we significantly reduce more than 90\% of the time cost, while keeping the performance of the protection.

\textbf{Anti Dynamic Compression}. Unlike saving images, saving videos requires a codec, which applies compression algorithms to balance the file size and the video quality. 
When it comes to videos on the Internet, it's often more suitable to utilize variable bitrate as opposed to constant bitrate. This is due to the fluctuation in available bandwidth and the diverse nature of video content, which demands a dynamic compression ratio applied to each frame to ensure optimal video streaming.
Therefore, it is essential to preserve the perturbation information $\delta$ throughout the dynamic compression process to guarantee the effectiveness of the protection.
For this, we introduce a simple solution to achieve anti-dynamic compression. Specifically, we transform the perturbation from the model input space $[-1,1]$, which is common in previous methods, to the pixel space $[-255, 255]$, and further quantize the values of the perturbation to ensure they have shorter bits. Such operations are the approximated simulation of the compression from the codec. By adding such a simulation into the protecting process, we can make the perturbation more robust and less sensitive against compression. Through the experiments, we find that our method  
does not introduce any significant time overhead while
increasing the video bitrate by about 8\%, which means the video contains more information.

\begin{algorithm}[t]
   \caption{\SysName Algorithm}
   \label{alg:prime}
\begin{algorithmic}[1]
\begin{small}
   \STATE {\bfseries Input:} Video $V=[x_1, x_2, \dots, x_n]$, target images $Q=\{\hat{x}_1, \hat{x}_2, \dots, \hat{x}_N\}$, models $\mathcal{E}$, $\mathcal{D}$, $\mathcal{U}$, $\mathcal{E}_\mathrm{img}$, diffusion steps $T$, perturbation budget $\epsilon$, optimization steps $K$
   \STATE $V' = []$
   \FOR{$i = 1 \to n$}
    \STATE Obtain $j_i$ based on Eq.\ref{eq:2}
    \STATE Obtain target feature set $\hat{F}$ for $\hat{x}_{j_i}$
    \STATE Initialize perturbation $\delta_0$
    \FOR{$k=1\to K$}
    \STATE Obtain feature set $F$ for $\prod_\epsilon (x_i + \delta_{k-1})$
    \STATE Calculate loss based on Eq.\ref{eq:4}
    \STATE Update $\delta_{k-1} \to \delta_{k}$
    \STATE Obtain $c^i_k$ based on Eq.\ref{eq:3}
    \IF{$c^i_k$ is converged}
    \STATE Append $x'_i = \prod_\epsilon (x_i + \delta_k)$ to $V'$
    \STATE Break
    \ENDIF
    \ENDFOR
   \ENDFOR
   \STATE Return $V'$
   \end{small}
\end{algorithmic}
\end{algorithm}

\textbf{The Proposed \SysName}.
As shown in Algorithm~\ref{alg:prime}, we first obtain the most suitable target image for the current frame $x_i$, based on our fast convergence search method. After obtaining the target image $\hat{x}_{j_i}$, we consider calculating the features $\hat{F}$ for it. In Photoguard~\cite{salman_raising_2023-1}, only the final outputs from $\mathcal{D}$ are considered as the features. However, we find that during the diffusion forward process and the sampling process, the intermediate results are equally important because we do not have information about the number of diffusion steps $T$ used by the attacker. Disrupting intermediate results can improve the transferability of the perturbation during the diffusion process. On the other hand, calculating the gradient on the intermediate results will not bring additional computing costs, which makes it practical during the protection. To make $\mathcal{U}$ have contributions in the forward process, we adopt the DDIM inversion method~\cite{geyer_tokenflow_2024} to predict the noise with $\mathcal{U}$ to replace the original forward process. Therefore, in \SysName, we consider four sources of the features, i.e., features $f_{\mathcal{E}, 0}$ from $\mathcal{E}$, features $f^d_{\mathcal{E},t}$ from the DDIM inversion at time step $t$, features $f^s_{\mathcal{E},t}$ from the sampling process at time step $t$, and outputs $f_\mathcal{D}$ from $\mathcal{D}$. Especially, in the diffusion process, the prompt condition $c_p$ is empty, as we have no information about the editing prompt used by the attackers. Therefore, $\hat{F}$ can be written as
\begin{align*}
    \hat{F} = \{\hat{f}_{\mathcal{E}, 0}, \hat{f}^d_{\mathcal{E},1}, \dots, \hat{f}^d_{\mathcal{E},T}, \hat{f}^s_{\mathcal{E},T}, \dots, \hat{f}^s_{\mathcal{E},1},  \hat{f}_\mathcal{D} \},
\end{align*}
which is calculated on the target image $\hat{x}$. Specifically, when we use $T$ steps in the forward process and the sampling process, there will be $T$ features for $f^d_{\mathcal{E},t}$ and $f^s_{\mathcal{E},t}$, respectively. We consider all of them when computing the loss functions.

During the optimization process, we adopt the anti dynamic compression method to add the perturbation $\delta_k$ to the clean frame $x_i$, which is represented by $\prod_\epsilon (x_i + \delta_{k})$ under the budget $\epsilon$. Similarly, we compute the features $F$ for the perturbed input $\prod_\epsilon (x_i + \delta_{k})$. Then we compute the loss:
\begin{align}\label{eq:4}
    L &= \vert f_{\mathcal{E}, 0} - \hat{f}_{\mathcal{E}, 0}\vert_1 + \vert f_\mathcal{D} - \hat{f}_\mathcal{D}\vert_1 \\ \nonumber
    &+ \sum_{t=1}^T (\vert f^d_{\mathcal{E},t} - \hat{f}^d_{\mathcal{E},t} \vert_1 + \vert f^s_{\mathcal{E},t} - \hat{f}^s_{\mathcal{E},t} \vert_1), 
\end{align}
where $\vert \cdot \vert_1$ is the $L_1$-norm. To update the perturbation $\delta_k$, we minimize $L$, i.e., $\min_\delta L$, which gives results closer to the target image $\hat{x}_{j_i}$. After updating $\delta$, we examine the convergence of the optimization process, by computing $c^i_k$ (Eq.\ref{eq:3}). If the optimization process is converged at step $k$, we will use $\prod_\epsilon (x_i + \delta_k)$ as $x_i'$ and start to optimize the next frame. Otherwise, we will continue to optimize the current frame till convergence or reaching the attack budget.

Overall, \SysName combines our new proposed mechanisms to accelerate the optimization process and restore more information for the compressed videos. Furthermore, these mechanisms are general and not tailor-made for \SysName, which means future works can directly adopt them to enhance their own performance.

\section{Experiments}
\label{sec:exp}

\begin{table*}[h]
\centering
\begin{adjustbox}{max width=1.0\linewidth}
\begin{tabular}{c|c|c|c|c|c|c|c|c|c|c|c}
 \Xhline{2pt}
\textbf{Name} & Donald Trump & Drake & Joe Biden & Katy Perry & Messi & Rihanna & Robert Downey Jr. & Ryan Gosling & Scarlett Johansson & Taylor Swift & \textbf{Sum} \\ \hline
\textbf{\# of clips} & 6 & 2 & 4 & 6 & 2 & 2 & 2 & 3 & 2 & 6 & 35 \\ \hline
\textbf{\# of total frames} & 1027 & 141 & 606 & 574 & 207 & 258 & 285 & 325 & 263 & 430 & 4116 \\
 \Xhline{2pt}
\end{tabular}
\end{adjustbox}
\caption{Details of \DataName. It contains 10 famous people and 35 video clips in total.}
\vspace{-10pt}
\label{tab:dataset}
\end{table*}

\subsection{Data Collection}
\label{sec:dataset}

We notice that no public standard benchmark and dataset have been previously proposed for malicious purposes in the video editing task. To evaluate our protection method and compare it with baselines, we build a dataset, \textbf{VI}deos f\textbf{O}r ma\textbf{L}icious \textbf{E}diting a\textbf{N}d pro\textbf{T}ection, \DataName.


To simplify the data collection process and facilitate subsequent evaluation stages, we choose to collect videos of famous people, such as actors and politicians, from various sources on the Internet.
Initially, we conduct a manual assessment to identify celebrities and politicians who can be perfectly and realistically generated by the existing LDMs. 
To ensure diversity in terms of gender, age, and race, we carefully select individuals.
After determining the list of celebrities and politicians, we proceed to acquire their videos from the Internet. 
Most of these videos are sourced from official channels, while the remaining content is gathered from public channels.
\textbf{\textit{Due to copyright issues, we are unable to make this collected dataset publicly available}.}

Upon obtaining the videos, we manually edit and cut them into scene-consistent video clips, each comprising tens to hundreds of frames. 
We carefully filter out clips that contain transitions, illumination changes, or main object changes.
%
Subsequently, we create specific configurations for malicious editing for each clip. For each configuration, we tune the malicious prompts and adjust other hyperparameters used in the video editing pipelines for different LDMs. This entire process requires hundreds of GPU hours of effort.

We summarize the details of \DataName, in Table~\ref{tab:dataset}. There are 10 identities in \DataName with 6 males and 4 females. They can be grouped into politicians (i.e., Donald Trump and Joe Biden), singers (i.e., Drake, Katy Perry, Rihanna, and Taylor Swift), actors (i.e., Robert Downey Jr., Ryan Gosling, and Scarlett Johansson), and athletes (i.e., Messi). Specifically, all original videos we collect are in resolution $1280\times 720$, which is the most popular format on the Internet. 

To create prompts for malicious editing tasks, we follow a very simple and direct template: ``[Someone] [Do Something] [Somewhere]". We first create a description for each original video based on this template. For the malicious NSFW editing task, we keep the ``[Someone]" part and change the ``[Do Something]" and ``[Somewhere]". For example, if we want to generate a video in which the person is naked, we will replace ``[Do Something]" with ``is naked" or ``is nude". If we want to generate a bloody video, we will replace ``[Somewhere]" with ``in a bloody scene". For the malicious swapping editing task, we only replace ``[Someone]" with a new name. For example, if we want to generate a video of Donald Trump based on a video of Joe Biden, we will replace ``Joe Biden" with ``Donald Trump". Furthermore, we adopt the prompt weighting method, Compel\footnote{https://github.com/damian0815/compel}, to manually adjust different weights for ``[Someone]", ``[Do Something]", and ``[Somewhere]" to obtain the best editing results. In total, we have designed 280 attacking configurations,
which will be used for evaluating the generated videos in subsequent experiments.


\subsection{Experiment Settings}

\textbf{Editing Models}. In our experiments, we consider public LDMs for high-quality and realistic video generation and editing. After evaluating accessible models on the Hugging Face, we manually select four different models, i.e., Stable Diffusion v1-5 (SD1.5)\footnote{https://huggingface.co/runwayml/stable-diffusion-v1-5}, Dreamlike Photoreal 2.0 (DP)\footnote{https://huggingface.co/dreamlike-art/dreamlike-photoreal-2.0}, HassanBlend1.4 (HB)\footnote{https://huggingface.co/hassanblend/hassanblend1.4}, and RealisticVisionV3.0 (RV)\footnote{https://huggingface.co/SG161222/Realistic\_Vision\_V3.0\_VAE}. These four representative models have different advantages in generating realistic photos, under different conditions, such as various illumination and human poses. And our method can be generalized to other models as well.

\textbf{Editing Pipelines}. As we propose two types of malicious editing, we choose the most recent and representative open-source editing methods for them, respectively. Specifically, we experimentally find that TokenFlow~\cite{geyer_tokenflow_2024} is suitable for the malicious NSFW editing task, and Rerender A Video~\cite{yang_rerender_2023} is suitable for the malicious swap editing task. And our method can be generalized to other pipelines.

\textbf{Protection Settings}. In our threat model, the protectors have no information about the editing models and pipelines used by the malicious users. Therefore, we consider that the protectors generate perturbation based on a public model, and the performance of the perturbation is based on transferability. We adopt two public models, i.e., Stable Diffusion v1-5 (SD1.5) and Stable Diffusion v2-1 (SD2.1)\footnote{https://huggingface.co/stabilityai/stable-diffusion-2-1-base}. 

For the optimization details, we set the maximum number of optimization steps $K$ as 100, and the maximum perturbation size $\epsilon$ as 8 under $l_\infty$-norm for \SysName and Photoguard~\cite{salman_raising_2023-1}. On the other hand, we notice that Photoguard uses 4 steps in the diffusion sampling process. Therefore, to maintain the same computing budget, we use $T=2$ steps in the DDIM inversion and $T=2$ steps in the sampling process. 

To form the queue of target images, we use the validation set of ImageNet~\cite{deng_imagenet_2009} for \SysName. For Photoguard, we use its officially provided target image. The perturbed videos are saved using libx264 codec with variable bitrate and best quality preference. We resize the resolution of videos into $672\times 384$. For each video, we only use the first 40 frames to edit.

\subsection{Metrics}

\textbf{Subjective Metrics}. We consider six dimensions of human perception, i.e., \textit{Content Consistency}, \textit{Prompt Matching}, \textit{Naturalness}, \textit{Frame Stability}, \textit{Video Quality}, and \textit{Personal Preference}, which are aligned with a recent benchmark, VBench~\cite{huang_vbench_2023}. Specifically, \textit{Content Consistency} describes the extent to which the edited video maintains the same details as the original video, such as layout, style, movements, and expressions. \textit{Prompt Matching} describes how well the edited video matches the theme of the given prompt, e.g., sex and blood. \textit{Naturalness} describes how plausible the edited video looks. \textit{Frame Stability} measures the temporal consistency of the edited video. \textit{Video Quality} describes the overall quality of the edited video. All these metrics are normalized into the interval $[1, 5]$. Higher scores mean better performance. For \textit{Personal Preference}, we give interviewees several videos and ask them to pick the one they prefer.

\textbf{Objective Metrics}. For objective metrics, we consider peak signal-to-noise ratio (PSNR), structural similarity (SSIM), and VCLIPSim. Specifically, PSNR compares the noise ratio in the edited videos w/ and w/o protection. SSIM compares the perceived quality for the edited videos w/ and w/o protection. When computing PSNR and SSIM, we use the original unprotected videos to generate edited videos as references. VCLIPSim is supported by ViCLIP~\cite{wang_internvideo_2022}, which reflects the similarity between the video and the given prompt.

\begin{table*}[h]
\centering
\begin{adjustbox}{max width=1.0\linewidth}
\begin{tabular}{c|cccc|cccc}
 \Xhline{2pt}
\multirow{3}{*}{\textbf{Task}} & \multicolumn{4}{c|}{\textbf{SD1.5}} & \multicolumn{4}{c}{\textbf{SD2.1}} \\ \cline{2-9} 
 & \multicolumn{2}{c|}{Photoguard} & \multicolumn{2}{c|}{\SysName} & \multicolumn{2}{c|}{Photoguard} & \multicolumn{2}{c}{\SysName} \\ \cline{2-9} 
 & \textbf{PSNR $\downarrow$} & \multicolumn{1}{c|}{\textbf{SSIM $\downarrow$}} & \textbf{PSNR $\downarrow$} & \textbf{SSIM $\downarrow$} & \textbf{PSNR $\downarrow$} & \multicolumn{1}{c|}{\textbf{SSIM $\downarrow$}} & \textbf{PSNR $\downarrow$} & \textbf{SSIM $\downarrow$} \\ \hline
\textbf{\begin{tabular}[c]{@{}c@{}}Malicious NSFW Editing\end{tabular}} & \textbf{18.30} & \multicolumn{1}{c|}{0.64} & {\ul 18.32} & \textbf{0.57} & 18.58 & \multicolumn{1}{c|}{0.65} & 18.42 & \textbf{0.57} \\ \hline
\textbf{\begin{tabular}[c]{@{}c@{}}Malicious Swapping Editing\end{tabular}} & 17.27 & \multicolumn{1}{c|}{0.63} & \textbf{16.93} & \textbf{0.62} & 17.39 & \multicolumn{1}{c|}{0.63} & {\ul 16.94} & \textbf{0.62} \\
 \Xhline{2pt}
\end{tabular}
\end{adjustbox}
\vspace{-5pt}
\caption{PSNR and SSIM under different protection methods. 
\textbf{Bold} for the best results and {\ul underline} for the second-best results.}
\vspace{-15pt}
\label{tab:psnr}
\end{table*}

\begin{table}[h]
\centering
\begin{adjustbox}{max width=1.0\linewidth}
\begin{tabular}{c|c|cccc}
 \Xhline{2pt}
\multirow{3}{*}{\textbf{Task}} & \multirow{2}{*}{\textbf{Original}} & \multicolumn{2}{c|}{\textbf{SD1.5}} & \multicolumn{2}{c}{\textbf{SD2.1}} \\ \cline{3-6} 
 &  & \multicolumn{1}{c|}{Photoguard} & \multicolumn{1}{c|}{\SysName} & \multicolumn{1}{c|}{Photoguard} & \SysName \\ \cline{2-6} 
 & \textbf{VCLIPSim} & \multicolumn{4}{c}{\textbf{VCLIPSim $\downarrow$}} \\ \hline
\textbf{\begin{tabular}[c]{@{}c@{}}Malicious\\ NSFW Editing\end{tabular}} & 0.2251 & \multicolumn{1}{c|}{0.2151} & \multicolumn{1}{c|}{\textbf{0.2025}} & \multicolumn{1}{c|}{0.2160} & {\ul 0.2028} \\ \hline
\textbf{\begin{tabular}[c]{@{}c@{}}Malicious\\ Swap Editing\end{tabular}} & 0.2076 & \multicolumn{1}{c|}{0.2030} & \multicolumn{1}{c|}{\textbf{0.2022}} & \multicolumn{1}{c|}{0.2034} & {\ul 0.2027} \\
 \Xhline{2pt}
\end{tabular}
\end{adjustbox}
\vspace{-5pt}
\caption{VCLIPSim under different protection methods. 
\textbf{Bold} for the best results and {\ul underline} for the second-best results.}
\vspace{-5pt}
\label{tab:clip}
\end{table}

\begin{figure*}[ht]
\centering
\begin{subfigure}[b]{0.45\linewidth}
\centering
\includegraphics[width=\linewidth]{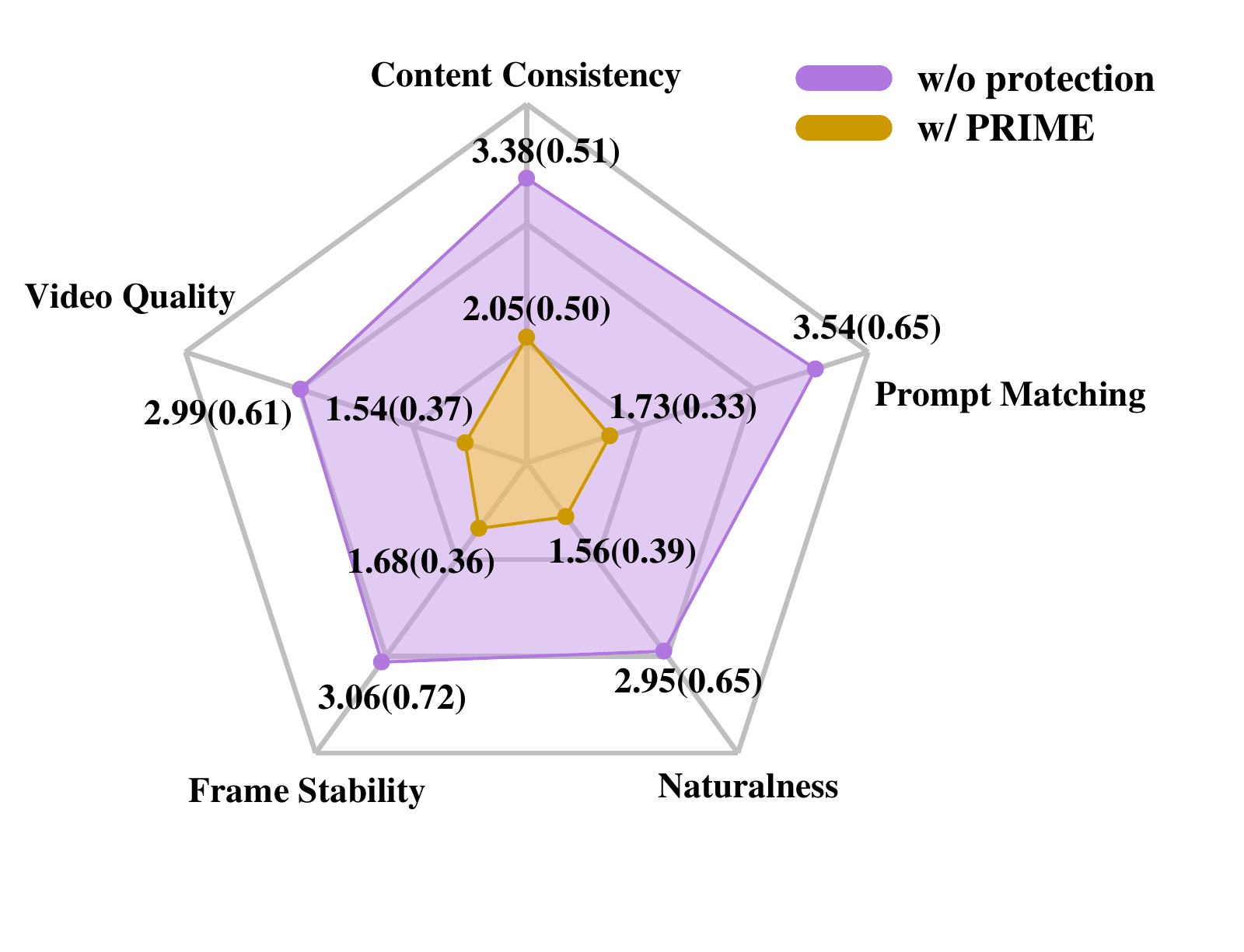}
\vspace{-38pt}
\caption{Evaluation results for malicious NSFW editing.}
\label{fig:nsfw} 
\end{subfigure}
\begin{subfigure}[b]{0.45\linewidth}
\centering
\includegraphics[width=\linewidth]{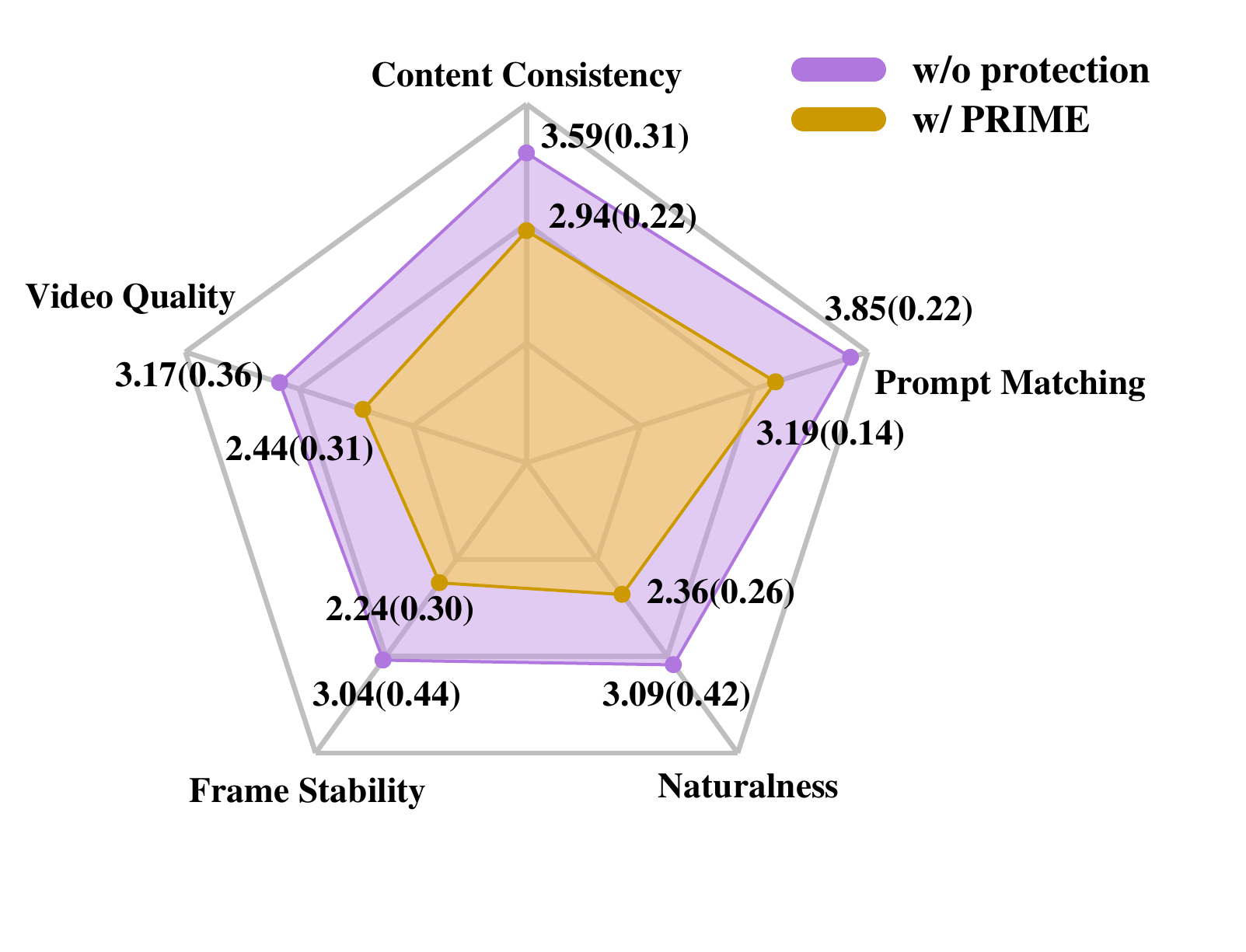}
\vspace{-38pt}
\caption{Evaluation results for malicious swapping editing.}
\label{fig:swap} 
\end{subfigure}
\vspace{-10pt}
\caption{Human evaluation on six subjective metrics for two malicious editing tasks. We show the mean score for each metric and put the standard deviation in brackets.
The results are calculated on the edited videos.
}
\label{fig:subjective} 
\vspace{-10pt}
\end{figure*}

\begin{table}[h]
\centering
\begin{adjustbox}{max width=1.0\linewidth}
\begin{tabular}{c|c|c|c}
 \Xhline{2pt}
\textbf{Task} & Photoguard* & Photoguard & \SysName \\ \hline
\textbf{Malicious NSFW Editing} & 35\% & 47\% & \textbf{18\%} \\ \hline
\textbf{Malicious Swap Editing} & 49\% & 42\% & \textbf{9\%} \\ 
 \Xhline{2pt}
\end{tabular}
\end{adjustbox}
\vspace{-5pt}
\caption{Human preference study. A lower probability is better for the protection method. \textbf{Bold} for the best results. We provided respondents with three video clips protected by different methods and asked them to choose the clip that looks best. Photoguard* stands for the Encoder attack.}
\vspace{-17pt}
\label{tab:vote}
\end{table}

\subsection{Protection Evaluation}

\textbf{Time Consumption and Bitrate Comparison}. 
We compare the time consumption and bitrate to intuitively validate the effectiveness of our proposed mechanisms in terms of time-saving and resistance to dynamic compression. 
We record the time cost on one NVIDIA RTX A6000. 
Notably, both Photoguard and \SysName have a maximum batch size limit of 1 due to GPU memory constraints.
For the average time consumption for protecting a video containing 40 frames, Photoguard will cost 20,500 seconds. \SysName will only cost 1,700 seconds, which is only 8.3\% of the cost of Photoguard. For the GPU memory occupation, Photoguard takes about 17GB and \SysName takes about 20GB. The additional occupation is because of the CLIP model used in searching the target images. Overall, both methods can run on consumer-grade GPUs, such as the RTX 4090.
Furthermore, the saved protected videos from Photoguard have an average bitrate of 45021 kbps. These from \SysName have an average bitrate of 45776 kbps, which is 8\% higher than those from Photoguard.
Overall, the results prove that our proposed methods collaborating with \SysName can significantly reduce the time consumption and improve the bitrate, keeping more information in the saved videos.

\textbf{Based on Objective Metrics}. 
We adopt the generated videos based on videos without protection as references, and compute PSNR and SSIM for the generated videos based on videos with protection. In Table~\ref{tab:psnr}, we show the protection results on different source models, i.e., SD1.5 and SD2.1, to compare the transferability of different protection methods. The results indicate that even though \SysName costs less time, it achieves better protection results on the pixel level. Furthermore, \SysName shows better transferability when we compare the results on different editing tasks and source models.

We compare the prompt consistency for edited videos in Table~\ref{tab:clip}. The results indicate that \SysName has better performance in breaking the connections between edited frames and the prompt and has better transferability between models and pipelines. We notice that although the ViCLIP is trained on a large video dataset, it may not be very suitable for evaluating malicious videos. The main reason is that the ViCLIP training set is filtered and contains only a small part of NSFW content and famous people. It will be more reasonable to adopt a fine-tuned ViCLIP for future work. But it is out of our paper's scope. We appeal that we should use human evaluation as the primary evaluation reference.

\textbf{Based on Subjective Metrics}. We evaluate the edited videos based on subjective metrics because the generated videos mainly aim to mislead humans and cause negative impacts. To obtain the subjective metrics, we invited interviewees to complete questionnaires. We strictly protect the privacy of all interviewees, without collecting any of their personal information. Considering the cost of evaluating with human interviewees, we provide the quality results for \SysName and provide human preference results for Photoguard and \SysName. After collecting 140 results for each, we present the evaluation results in Figure~\ref{fig:subjective} and Table~\ref{tab:vote}.

\vspace{-0.5em}
Based on the quality results in Figure~\ref{fig:subjective}, we have two conclusions: \ding{182} Existing video editing pipelines can successfully execute the malicious editing tasks and achieve high \textit{Prompt Matching} scores (3.54 out of 5 and 3.85 out of 5) and high \textit{Video Quality} scores (2.99 out of 5 and 3.17 out of 5); \ding{183} \SysName can effectively protect videos by decreasing \textit{Prompt Matching} scores (1.73 out of 5 and 3.19 out of 5) and \textit{Video Quality} scores (1.54 out of 5 and 2.44 out of 5). 

\vspace{-0.5em}
With \SysName, human faces are disrupted, and edited frames are less affected by the given prompts. Therefore, \textit{Prompt Matching} scores decrease in both tasks. On the other hand, the stability of the frame and the consistency of the video decrease. There are many flickers between frames, making the video less natural. We can obtain the aforementioned conclusion based on \textit{Content Consistency} scores (from 3.38 to 2.05 and from 3.59 to 2.94), \textit{Frame Stability} scores (from 3.06 to 1.68 and from 3.04 to 2.24), and \textit{Naturalness} scores (from 2.95 to 1.56 and from 3.09 to 2.36).

On the other hand, for our proposed two types of malicious editing, we find that changing the identities of the original videos is easier than adding blood or naked bodies to the original videos, and protecting videos from NSFW editing is easier and more effective. The reasons are mainly from two aspects. First, the swap task is easier because it keeps the layout and other elements unchanged in the videos. The NSFW editing task requires changing the layout and pixel-level details, which is closely related to the capability of LDMs to understand the prompt correctly. Second, for the swap task, we adopt additional constraints in the pipeline, such as the optical flow model~\cite{xu_gmflow_2022} and the canny edge model~\cite{zhang_adding_2023}, which enhance the stability and robustness of the edited videos. For NSFW editing, we only use global attention and cross-frame attention to keep the frame consistency because we find that the additional constraints mainly influence the layout and pixel-level details, which conflict with the target of the NSFW editing and makes the video editing pipeline give unchanged results for NSFW prompts.

\vspace{-0.5em}
Besides these subjective metrics, we ask the respondents to select videos with better quality among three edited videos, which are generated from videos protected by Photoguard*, Photoguard, and \SysName. The results in Table~\ref{tab:vote} prove that \SysName will better decrease the quality of the generated videos. More than 80\% of the interviewees think that the generated videos from videos protected by Photoguard have better quality. Therefore, \SysName is more effective in protecting videos from malicious editing.

\vspace{-0.5em}
Overall, \SysName achieves better protection performance than Photoguard and reduces the overall time cost. \textbf{\textit{The visualization results can only be obtained by contacting us, due to the legal considerations}.}

\section{Conclusion}
\vspace{-0.5em}

In this paper, we study malicious editing tasks in the video area. Through the comprehensive exploration, we find it is a huge threat to the public, as malicious people can easily use video editing pipelines to conduct two types of malicious editing tasks. To better protect the videos from malicious editing, we propose a new method, \SysName, which can significantly reduce the time cost for video protection and improve the bitrate for the protected videos. Based on our evaluation, we prove the advantages of \SysName in protecting videos from editing. We believe that our work will promote future work in protecting videos and portrait rights.

\clearpage

\section*{Social Impacts, Ethical and Legal Concerns}

It has been a very long story to protect portrait rights and other intellectual property since the generative models became popular on the Internet. For example, a very recent case is the AI-generated Taylor Swift photos circulating on social networks, which raises the attention of many popular presses, such as BBC, and even the White House. Not just for famous people, advanced generative models can affect normal people's photos as well, due to editing technologies and personalization technologies. Such powerful generative models can be used for illegal use. Therefore, adding watermarks into the generative models to detect AI-generated content, adding perturbation to images to immunize editing, and detecting illegal content before returning the results to users are three main solutions in addressing the ethical and legal concerns of using generative AIs.

Similar to generating images or editing images, generating videos and editing videos become possible with advanced generative AI models. Although it is still at the beginning phase, we find that the rising video editing pipelines equipped with advanced latent diffusion models are able to do some operations, such as swapping faces, adding naked bodies, and replacing backgrounds. This finding motivates us to study the risks for the public under the threat of such video editing operations.

First, at the forefront of these concerns is collecting other people’s selfie videos is easy. Sharing selfie videos to social applications, such as TikTok and Instagram, is very popular among Gen Z (people who are born between the mid-to-late 1990s and the early 2010s). People's desire to share becomes an exact threat to themselves. Because others can easily download their videos from social applications and edit them. Based on this point, social applications should protect their users' shared images and videos.

The second concern is from an ethical standpoint for these AI-generated videos and AI-edited videos, which pose a significant threat to the veracity of information and encourage the dissemination of misinformation, defamation, and deleterious content. These influences jeopardize the integrity of information dissemination, eroding the public's trust in online platforms, such as the media and social applications, as reliable sources of information. Furthermore, malicious editing may contravene privacy statutes, transgress intellectual property safeguards, and incur charges of defamation. This complex legal terrain demands a nuanced examination of liability, accountability, and the potential gaps in existing legal frameworks that may be exploited by malicious actors. Especially, the existing legal system has not provided enough evidence and support for the judges to determine a potential crime.

The third concern is that platforms that host user-generated content can promote and encourage users to create these AI-generated videos and AI-edited videos. These platforms should provide users rights to generate and edit videos and supervise users' activities to avoid providing potentially illegal or harmful videos. We find that some existing platforms, such as Gen-1 (\url{https://research.runwayml.com/gen1}), provide a detection model to detect whether the generated video contains illegal or harmful content. However, other platforms, such as Pika (\url{https://pika.art/})\footnote{We tested the Beta test version.}, do not detect the outputs, making it possible to generate videos containing naked bodies and blood.

In addition, the issues caused by the illegal and harmful videos generated by AI encompass broader societal implications due to the Internet. For example, the unchecked proliferation of illegal and harmful videos can sow discord, amplify existing divisions, and potentially incite harm between people. It is possible to create AI-generated videos, such as an imitation of the Murder of George Floyd, causing arguments on the Internet between different races.

Due to the flaws and imperfections in the current legal system and social framework, we call on users to protect their videos and legitimate rights and interests. We propose a protection method to help users who are willing to share their selfie videos on social media to protect these videos. With our protection, it will be easier for humans to recognize AI-edited videos. In this way, we hope that our approach can provide an interim solution until a full legal and regulatory system is proposed.

\bibliography{bib}
\bibliographystyle{icml2024}

\newpage
\appendix
\onecolumn

\section{Hyperparameters}

We introduce the hyperparameters used in the two video editing pipelines, i.e., Tokenflow and Rerender A Video, as well as the prompt weights. For Tokenflow, the number of DDIM inversion steps is 500, which is its default setting. The guidance scale is 7.5. Other parameters related to the injections are their default values without changing. For Rerender A Video, the interval is set to 1, due to the limited number of frames. The strength of the first frame is 0.8 and the strength of the ControlNet is 0.7. We manually search for these two parameters, which is better than other combinations. The ControlNet we use is the Canny Edge, which we find is better than HED Boundary and Depth. We also set the loose attention, considering the motions in the videos, to obtain a better result. Other hyperparameters are default. We use an additional prompt ``RAW photo, subject, (high detailed skin:1.2), 8k uhd, dslr, soft lighting, high quality, film grain, Fujifilm XT3'' as a supplement for the prompt.

For both pipelines, the negative prompt is set as ``deformed iris, deformed pupils, semi-realistic, cgi, 3d, render, sketch, cartoon, drawing, anime, mutated hands and fingers, deformed, distorted, disfigured, poorly drawn, bad anatomy, wrong anatomy, extra limb, missing limb, floating limbs, disconnected limbs, mutation, mutated, ugly, disgusting, amputation''. For Tokenflow, the prompt weights are in the interval of $[1.0, 1.61]$. For Rerender A Video, the prompt weights are in the interval of $[1.0, 1.4]$.

\section{Visualization Results}

Sorry for the inconvenience that we cannot provide visualization results with a public link, due to the reasonable concern that the edited content could be leaked to the Internet and cause unnecessary harm to these innocent people. \textbf{Therefore, the visualization results can only be obtained by contacting us, due to the legal considerations.} To further promise that the edited content does not cause further negative impacts, we add visible watermarks and limit the visualization similarity of the faces.

\section{Accessible Resources for \DataName}

Although it could be illegal to release the whole dataset, we consider only providing the configuration files used in our experiments. These configurations can be obtained by contacting us. For each model we use in the experiments, we tune the prompts to achieve better editing results. Even though we cannot provide the source videos to the public, protecting these people's rights, these configurations could be useful for future studies. We believe that the readers can obtain enough information about the source videos from these configurations because we describe each video clip we use. On the other hand, for other researchers, who are going to study in this area, we are willing to guide how to construct the dataset after signing a commitment. Note that in whatever case, we will not leak the source video clips and re-distribute them.


\end{document}